%% file: main.tex
\documentclass[runningheads]{llncs}
\usepackage{array}
\usepackage{color}

\usepackage{algorithm}% let algorithm use the patched version of the float package
\usepackage{algorithmic}
\usepackage{amsmath,amssymb,amsfonts}
\usepackage{graphicx,tabularx,booktabs}

\input{01_Preamble/CommonPackages.tex}

\definecolor[named]{myColorMainA}{RGB}{0,26,153}
\definecolor[named]{myColorMainB}{RGB}{174,49,54}

\newcommand{\xV}{\mathbf{x}}

\begin{document}
\title{Complex-valued embeddings\\ of generic proximity data\thanks{\tiny{MM is supported by the ESF program WiT-HuB 4/2014-2020, project KI-trifft-KMU, StMBW-W-IX.4-6-190065. M.B. and M.S. acknowledge support through the Northern Netherlands Region
of Smart Factories (RoSF) consortium, lead by Noordelijke Ontwikkelings
en Investerings Maatschappij (NOM), The Netherlands, see http://www.rosf.nl
}}}
\author{Maximilian M\"unch\inst{1,2}\orcidID{0000-0002-2238-7870} \and
Michiel Straat\inst{2}\orcidID{0000-0002-3832-978X} \and
Michael Biehl\inst{2}\orcidID{0000-0001-5148-4568}
Frank-Michael Schleif\inst{1}\orcidID{0000-0002-7539-1283}
}

\authorrunning{Maximilian M\"unch et al.}
\institute{University of Applied Sciences W\"urzburg-Schweinfurt,\\
Department of Computer Science and Business Information Systems,\\
D-97074 W\"urzburg, Germany \email{\{maximilian.muench, frank-michael.schleif\}@fhws.de}
\and University of Groningen, Bernoulli Institute for Mathematics,\\ Computer Science and Artificial Intelligence, \\P.O. Box 407, NL-9700 AK Groningen, The Netherlands \email{\{m.biehl,m.j.c.straat@rug.nl\}@rug.nl}}

\maketitle              % typeset the header of the contribution
\begin{abstract}
Proximities are at the heart of almost all machine learning methods. If the input data are given as numerical vectors of equal lengths,
euclidean distance, or a Hilbertian inner product is frequently used
in modeling algorithms.
In a more generic view, objects are compared by a (symmetric) similarity or dissimilarity measure, which may not obey particular mathematical properties. 
This renders many machine learning methods invalid, leading to convergence problems and the loss of guarantees, like generalization bounds.
In many cases, the preferred dissimilarity measure is not metric, like the earth mover distance, or the similarity measure may not be a simple inner product in a Hilbert space but in its generalization a Krein space.
If the input data are non-vectorial, like text sequences, proximity-based learning is used or ngram embedding techniques can be applied. Standard embeddings lead to the desired fixed-length vector encoding, but are costly and have substantial limitations in preserving the original data's full information.
As an information preserving alternative, we propose a complex-valued vector embedding of proximity data.
This allows suitable machine learning algorithms to use these fixed-length, complex-valued vectors for further processing.
% The obtained fixed length, complex-valued, vectors can be used by suitable machine learning algorithms.
%
%It is shown how a complex-valued embedding can be efficiently determined for non-metric and non-psd proximity data.
The complex-valued data can serve as an input to complex-valued machine learning algorithms.
In particular, we address supervised learning and use extensions of prototype-based learning.
The proposed approach is evaluated on a variety of standard benchmarks and shows strong performance compared to traditional techniques in processing non-metric or non-psd proximity data.
\end{abstract}
\section{Introduction}
Machine learning has a growing impact in various fields and the considered input data become more and more generic \cite{Pekalska2005a,DBLP:journals/air/HalpinM13}. In particular non-vectorial data like
text data, biological sequence data, graphs and other input formats are used \cite{DBLP:journals/bigdata/Hendler14}.
The vast majority of learning algorithms are expecting fixed-length real value vector data as inputs and can not directly be used on non-standard data \cite{Pekalska2005a,DBLP:conf/sspr/PekalskaDGB04}.

Using embedding approaches is one strategy to obtain a vectorial embedding, but this is costly and information is only partially preserved \cite{Pekalska2005a}.
More recent approaches of deep learning like word2vec or others require the training of the embedding model and are only effective if the set of input data is large \cite{GoodBengCour16}.

In a more generic scenario, proximity measures, like alignment functions, can be applied to compare non-vectorial objects to obtain a proximity score between two objects. These values can be used in the modeling step.

If all input data are pairwise compared, we obtain a proximity matrix $P \in \mathbf{R}^{N \times N}$.
If the measure is a metric dissimilarity measure, we have a distance matrix that can be used for the nearest-mean classifier.
In case of inner  products like the euclidean inner product or the RBF-kernel function, a kernel matrix is obtained.
If this kernel matrix is positive semidefinite (psd), a large number of kernel methods can be used in the modeling stage \cite{Cristianini2004a}.

Also so-called empirical feature space approaches have been considered, but with the drawback of high model complexity and inherent data transformations \cite{DBLP:conf/nips/KarJ11}.
For a more in-depth introduction into indefinite learning see \cite{Schleif2015f,DBLP:journals/tkde/ChenL08}

Here we consider non-vectorial input data given either by a non-metric dissimilarity measure or a non-standard inner product, leading to an indefinite kernel function.
As detailed in \cite{Schleif2015f}, learning models can be calculated on these generic proximity data in very different ways.
Most often, the proximities are transformed to fit into classical machine learning algorithms, with a number of limitation \cite{Schleif2015f}.
In this work, we propose the application of a complex-valued embedding on these data.
Recently different classical learning algorithms have been extended to complex-valued inputs \cite{Gay2016}. %TODO citations
It is now possible to preserve the information provided in the generic proximity data while learning in a fixed-length vector space using
a highly effective, well-understood learning algorithm.
The respective procedures are detailed in the following and evaluated on classical benchmark data with strong results.

\section{Background and basic notation}\label{sec:related}
Consider a collection of $N$ objects $\mathbf{x}_i$, $i=\{1,2,...,N\}$,  in some input space $\mathcal{X}$. 
Given a similarity function or inner product on $\mathcal{X}$, corresponding to a metric, one can construct a proper Mercer kernel acting on pairs of points from  $\mathcal{X}$.
For example, if  $\mathcal{X}$ is a finite-dimensional vector space, a classical similarity function in this space is the  Euclidean inner product (corresponding to the Euclidean distance).
\subsection{Positive definite kernels - Hilbert space}
The Euclidean inner product is also known as linear kernel with $k(\mathbf{x},\mathbf{x^\prime})=\langle \phi(\mathbf{x}) , \phi(\mathbf{x^\prime}) \rangle$, 
where 
%$\langle \phi(\mathbf{x}) , \phi(\mathbf{x^\prime}) \rangle$ is the Euclidean inner product and 
$\phi$ is the identity mapping. Another prominent kernel function 
is the RBF kernel $k(\mathbf{x},\mathbf{x^\prime})=\exp{\left (-\frac{||\mathbf{x}-\mathbf{x^\prime}||^2}{2\sigma^2}\right)}$,
with $\sigma>0$ as a free scale parameter. 
In any case, it is always assumed that the kernel function $k(\mathbf{x},\mathbf{x^\prime})$ is psd. 

%Now, let  $\phi: \mathcal{X} \mapsto \mathcal{H}$ be a \emph{general} mapping of patterns from $\mathcal{X}$ to a Hilbert space $\mathcal{H}$ equipped with the inner product $\langle \cdot, \cdot \rangle_\mathcal{H}$.
The transformation $\phi$ is, in general, a \emph{non-linear} mapping to a high-dimensional Hilbert space $\mathcal{H}$ and may not be given in an explicit form, but allowing \emph{linear} techniques in $\mathcal{H}$. Instead of providing an explicit mapping, a kernel function $k: \mathcal{X} \times \mathcal{X}
\mapsto \mathbb{R}$ is given, which encodes the inner product in $\mathcal{H}$.
The kernel $k$ is a positive \mbox{(semi-)} definite function such that $k(\mathbf{x},\mathbf{x}^\prime)  =\langle \phi(\mathbf{x}), \phi(\mathbf{x}^\prime) \rangle_\mathcal{H}$,
 for any $\mathbf{x},\mathbf{x}^\prime \in \mathcal{X}$. 
The matrix $K_{i,j}:= k(\mathbf{x}_i,\mathbf{x}_j)$ is an $N \times N$ kernel (Gram) matrix derived from the training data.
For more general similarity measures, subsequently, we also use $\mathbf{S}$ to describe a similarity matrix.

Kernelized methods process the embedded data points in a feature space utilizing only the inner products $\langle \cdot,\cdot \rangle_{\mathcal{H}}$ 
 \cite{Cristianini2004a}, without the need to explicitly calculate $\phi$, known as \emph{kernel trick}.
 However, to employ the benefits of linear methods also 
 explicit mappings of psd kernel function, using
 random Fourier features, are frequently used \cite{DBLP:conf/nips/RahimiR07}. 

However, this assumption is not always fulfilled and the underlying similarity measure may not be metric and hence not lead to a Mercer kernel.
Examples can be easily found in domain-specific similarity measures, as mentioned before.
Such similarity measures imply \emph{indefinite} kernels, preventing standard "kernel-trick" methods developed for Mercer kernels to be applied. 

\subsection{Non-positive definite kernels - Krein space}
A Krein space is an \emph{indefinite} inner product space endowed with a Hilbertian topology.
%, however the  inner products can be negative.
Let $\mathcal{K}$ be a real vector space. 
%$\mathbb{R}$
% or $\mathbb{C}$ (here we will use $\mathbb{R}$). 
An inner product space with an indefinite inner product
$\langle \cdot, \cdot \rangle_\mathcal{K}$ on $\mathcal{K}$ is a bi-linear form where all $f,g,h \in \mathcal{K}$ and $\alpha \in \mathbb{R}$ obey the
following conditions:
\begin{itemize}
    \item Symmetry: $\langle f,g \rangle_\mathcal{K} = \langle g,f \rangle_\mathcal{K}$; 
    \item linearity: 
$\langle \alpha f + g,h \rangle_\mathcal{K} = \alpha \langle f,h \rangle_\mathcal{K} + \langle g,h \rangle_\mathcal{K}$;
    \item %$ \forall g \in \mathcal{K}$ with
$ \langle f, g \rangle_\mathcal{K} = 0$ implies $ f = 0$. 
\end{itemize}
An inner product is positive semidefinite if $\forall f \in \mathcal{K}$,  $ \langle f,  f \rangle_\mathcal{K} \ge 0$, negative definite if $\forall f \in \mathcal{K}$, $\langle f, f \rangle_\mathcal{K} < 0$, otherwise it is indefinite.
A vector space $\mathcal{K}$ with inner product $\langle \cdot, \cdot \rangle_\mathcal{K}$ is called an inner product space.

An inner product space $(\mathcal{K}, \langle \cdot, \cdot \rangle_\mathcal{K})$ is a Krein space if we have two Hilbert spaces $\mathcal{H}_+$
and $\mathcal{H}_-$ spanning $\mathcal{K}$ such that $\forall f \in \mathcal{K}$ we have $f = f_+ + f_-$ with $f_+ \in \mathcal{H}_+$
and $f_- \in \mathcal{H}_-$ and 
$\forall f,g \in \mathcal{K}$, $ \langle f,g \rangle_\mathcal{K} = \langle f_+,g_+ \rangle_{\mathcal{H}_+} - \langle f_-,g_- \rangle_{\mathcal{H}_-}$.
%??? The space $\mathbb{R}^{p+q}$ with the usual positive definite inner product
%$\langle x, x^\prime \rangle:= x^\top x^\prime$ is called
%the associated Euclidean space.

Indefinite kernels are typically observed by means of domain-specific
non-metric similarity functions (such as alignment functions used in biology \cite{citeulike:668527}), by
specific kernel functions - e.g., the Manhattan kernel $k({x},{x^\prime})= - ||{x}-{x^\prime}||_1$,
tangent distance kernel \cite{DBLP:conf/icpr/HaasdonkK02} or divergence measures, plugged into standard kernel
functions \cite{Cichocki20101532}. 
%Another source of non-psd kernels are noise artifacts on standard kernel functions \cite{Haasdonk2005482}.
A finite-dimensional Krein-space is a so-called pseudo-Euclidean space.

\section{Embedding for non-psd proximities}
Embedding of a proximity matrix into a vector space is not a new consideration, see e.g. \cite{IOSIFIDIS2016190}, but was possible so far only in case of psd kernel functions. 
Given a symmetric \emph{dissimilarity} matrix with zero diagonal, an embedding of the data in a pseudo-Euclidean vector space,
determined by the eigenvector decomposition of the associated similarity matrix $\mathbf{S}$, is always possible \cite{Goldfarb1984575}
\footnote{The associated similarity matrix can be obtained by double centering \cite{Pekalska2005a} of the dissimilarity matrix.
$\mathbf{S} = -\mathbf{J} \mathbf{D} \mathbf{J}/2$ with $\mathbf{J} = (\mathbf{I}-\mathbf{1}\mathbf{1}^\top/N)$,
identity matrix $\mathbf{I}$  and vector of ones $\mathbf{1}$.}.
Given the eigendecomposition of  $\mathbf{S} $,  $\mathbf{S} = \mathbf{U} \mathbf{\Lambda} \mathbf{U}^\top$,
we can compute the corresponding vectorial representation $\mathbf{V}$
in the pseudo-Euclidean space by
\begin{equation}
\mathbf{V} = \mathbf{U}_{p+q+z} \left|\mathbf{\Lambda}_{p+q+z}\right|^{1/2}
\label{eq:embedding}
\end{equation}
where $\mathbf{\Lambda}_{p+q+z}$ consists of $p$ positive, $q$ negative non-zero eigenvalues and $z$ zero eigenvalues.
 $\mathbf{U}_{p+q+z}$ consists of the corresponding eigenvectors.
 The triplet $(p,q,z)$ is also referred to  as the signature of the
pseudo-Euclidean space.
The crucial point in Eq. \eqref{eq:embedding} is the \textit{absolute} operator used in the embedding, which is also called a flip operation in the field of indefinite learning \cite{Schleif2015f}.
This effectively makes the representation of the data metric again and can have a substantial (potentially negative) impact on the data representation and modeling performance \cite{DBLP:conf/icpram/MunchRBS20}. 

%To make the representation metric (for dissimilarities) or psd (for similarities) is assumed to be helpful.
The transformation of dissimilarities to obey metric properties, or of similarities to be psd is in general expected to be useful. 
At least, it is technically useful because it permits to employ many mathematical concepts, as shown in \cite{IOSIFIDIS2016190}, not available otherwise.
We will go a step further and remove the absolute function from the embedding in Eq.  \eqref{eq:embedding} and obtain Eq.  \eqref{eq:embedding_new}.
Accordingly, the new embedding does not modify the data, in particular, an inner product of the embedded data reveals the input again.
\begin{equation}
\mathbf{V} = \mathbf{U}_{p+q+z} \mathbf{\Lambda}_{p+q+z}^{1/2}
\label{eq:embedding_new}
\end{equation}
Subsequently, we show how a number of mathematical operations can still be used to get an efficient machine learning model.
This comes with additional effort and (finally) an embedding of the data into a \emph{complex-valued} vector space
is obtained. 
The embedding in Eq. \eqref{eq:embedding} and Eq.   \eqref{eq:embedding_new} is straight forward but extremely costly. 

Already in \cite{IOSIFIDIS2016190}, this was addressed for psd kernels (only) by using the Nystr\"om approximation.
The approach in  \cite{IOSIFIDIS2016190} can not be used directly in our setting since the input is non-psd.

In our former work \cite{Schleif2015g}, we have shown that the Nystr\"om approximation remains valid for generic proximity data, in particular similarities that are non-psd.
This has been recently reconsidered by a simplified
proof in \cite{DBLP:conf/icml/OglicG19}.
Hence the Nystr\"om approximation becomes available also to approximate
a non-psd matrix.
In our work \cite{Schleif2015g}, we have further shown how the Nystr\"om approximation can also be used to have an approximated Double-Centering to deal with dissimilarity data. 
Our former work helps in two ways to permit the embedding of Eq. \eqref{eq:embedding_new} effectively:
\begin{enumerate}
	\item the input needs not to be a kernel but can also be a dissimilarity matrix
	\item the Nystr\"om matrix approximation can also be done for non-psd similarities which reduced the costs of the embedding
\end{enumerate}
In the Nystr\"om approximation, we have to specify the number of $m$ landmarks with $m \ll N$.
The landmarks can be selected for non-psd matrices randomly
or by kmeans++ as shown recently in \cite{DBLP:conf/icml/OglicG17,DBLP:conf/icml/OglicG19}.
Our efficient approach to get an approximated complex-valued, vectorial embedding of a non-psd matrix is shown in Algorithm \ref{alg1}.

\begin{algorithm}
\begin{algorithmic}
\STATE {\bf Embed\_proximities(${P},m$)}
\IF{P is dissimilarity}
\STATE ${Knm,Kmm}$ := ApproximatedDoubleCentering(${P},m$) using \cite{Schleif2015g} and kmeans++ 
\ELSE
\STATE ${Knm,Kmm}$ := Approximate(${P},m$) using \cite{Schleif2015g}
for similarities and kmeans++
\ENDIF
\STATE $[C, A]$ := eig(Kmm);  with eigenvectors $C$ and eigenvalues in $A$ (diagonal)
\STATE $W$   := diag(sqrt(1./diag(A)))*C'   \; modified Nystr\"om projection matrix
\STATE $M$   := W*Knm' \qquad \qquad \qquad \qquad  complex-valued embedding 
\STATE $K^*$ := M'*M \qquad \qquad \qquad \qquad \quad reconstruction
\RETURN ${M}$
\end{algorithmic}
	\caption{Approximated embedding of symmetric proximities}\label{alg1}
\end{algorithm}

In the first step of Algorithm \ref{alg1}, the input matrix is approximated using the Nystr\"om approximation
(potentially with an integrated double centering). This can be done with linear costs and with guaranteed approximation
bounds \cite{Schleif2015g,DBLP:conf/icml/OglicG19}.
Subsequently, we calculate the essential part of the embedding
function in Eq. \eqref{eq:embedding_new} combined with the projection matrix of the Nystr\"om approximation,
 by taking the square root of the (pseudo-) inverse of the eigenvalue decomposition of $Kmm$.
 This can be done with linear costs as shown in \cite{Schleif2015g}.
The vectorial embedding $M$ is finally done by mapping the rectangular Nystr\"om part Knm of the similarities to the projection matrix $W$\footnote{The step how the projection matrix $W*$ is calculated, is slightly related to so-called Landmark MDS as suggested in \cite{DBLP:conf/eccv/BelongieFCM02} for psd-only matrices.}. The embedding is complex-valued if the similarity matrix $K$ is non-psd and hence $A$ contains negative values. 

We now have an approximated complex-valued fixed-length vectorial embedding of the proximity data $P$ whereby
the respective reconstruction is exact if the rank of $P$ equals to the number of non-vanishing eigenvalues in $A$. 
Algorithm \ref{alg1} has a linear complexity as long as the number of landmarks $m \ll N$, which is in general the
case.
In contrast to many other methods \cite{Schleif2015a}, the embedding procedure has a straight forward out of sample extension.
The mapping in \ref{alg1} can be done for new points by evaluating the proximity function for the landmark point and using the respective projection function.

For the complex-valued embedding (so far) a quite limited number of machine learning algorithm is available, like the complex-valued support vector machine (cSVM) \cite{ZHANG2010944,6868310}, the complex-valued generalized learning vector quantization (cGMLVQ) \cite{Schleif2012c,Straat2019}, or a complex-valued neural network (cNN) \cite{guberman2016complex,DBLP:conf/iclr/TrabelsiBZSSSMR18,DBLP:journals/corr/abs-1905-12321}.
Further, a nearest neighbor (NN) classifier by employing a standard norm operator can be used. While cSVM, cGMLVQ, cNN are parametric methods, the NN classifier is parameter-free and can be used directly.
In particular, after applying the norm, the obtained dissimilarity values are metric.
% one may add a proof here TODO (-> see definition of the norm 
For the cGMLVQ, we briefly recap the respective derivations and show how it can be effectively used within our setting. 

\section{Complex-valued Generalized Learning Vector Quantization}
\label{sec:cglvq}
In this section, we will review the \textit{complex-valued Learning Vector Quantization} (cGLVQ).
We first lay out the general idea of LVQ classification and the learning rules in modern variants.
We will then review the learning rules' adaptation to make the algorithm suitable for handling complex-valued data.

In Learning Vector Quantization (LVQ), the classification scheme is parameterized by a set of labeled prototypes and a distance measure $d(\cdot,\cdot)$.
New data is classified according to the label of the nearest prototype with respect to the distance measure $d(\cdot, \cdot)$.
This original idea was proposed by Kohonen in 1986.
In contrast to the $k$-nearest neighbor classifier in which the full dataset is used in the classification procedure, the classes in LVQ schemes are represented by only very few prototypes.
Hence, in the algorithm's working phase, LVQ takes less computational effort and storage in order to assign class labels to new data. Moreover, LVQ is often praised for its white-box character:
The prototypes that are crucial for the classification represent typical examples of the classes in the dataspace.
Numerous extensions of LVQ that have been proposed over the years, such as adaptive distance measures known as \textit{relevance learning}, low-rank approaches a.s.o, further increasing the performance and interpretability of the method.
The interpretability of LVQ provides important insights in many applications, see for instance \cite{VANVEEN2020105708,ArltBiehl2011} for examples in the medical domain. 

\subsection{Training an LVQ classifier}
Given a training dataset of $N$ labeled objects in $K$ classes, i.e. $(\mathbf{x}_i, y_i)$ with $i=\{1,2,...,N\}$, in which $\mathbf{x}_i \in \mathcal{X}$ is an input object and $y_i \in \{1,2,...,K\}$ its class label. The aim of the training procedure is choosing prototypes $\{(\mathbf{w}_k, y_k) \, | \, 1 \leq k \leq M\}$ in the space of the objects $\mathcal{X}$, such that the resulting classification scheme gives high classification accuracy with respect to unseen data.
The number of prototypes per class is a hyperparameter chosen by the user and one has to choose at least one prototype per class, hence the total number of prototypes $M$ is at least the number of classes $K$, i.e. $K \geq M$.
The distance measure $d(\cdot,\cdot)$ is of central importance in the training- and classification procedure.
It is used to compare input objects with prototype vectors for classification and for determining the change in positions of the prototypes in the training procedure.
If the data space $\mathcal{X}$ is Euclidean, the squared Euclidean distance measure is a common choice:
\begin{equation} \label{eq:eucliddist}
d^{\Lambda}(\mathbf{w}, \xV_i) = (\xV_i - \mathbf{w})^H \mathbf{\Omega}^H \mathbf{\Omega} (\xV_i - \mathbf{w}) \, ,
\end{equation}
in which $\mathbf{\Omega}$ is the identity matrix and $H$ is the Hermitian transpose. One drawback of this choice is that it is affected by differing scales of the input variables $x_i$, such that an $x_i$ with a large standard deviation compared to others has a large influence in the distance measure. This is, however, easily prevented by normalizing all input variables to unit standard deviation. A major improvement in LVQ is the introduction of \textit{relevance learning} \cite{Hammer2002,Schneider2009}, in which the elements of the linear projection $\mathbf{\Omega}$ are adapted during training to reflect the importance of the features in the classification task and to account for correlations between features.

The original idea of Kohonen's LVQ1 is to update the prototypes based on a heuristic Winner Takes All (WTA) rule. At the random presentation of an example, its closest prototype is attracted by the example if the class labels coincide and is repelled from the example otherwise.
In \cite{SatoYamada1995}, the authors proved that this training scheme does not converge and instead the authors propose a by now popular and successful cost function that does converge in the gradient descent.
The cost for an example is defined as:
\begin{equation} \label{eq:mu}
E_{GLVQ} = \sum_{i=1}^P \Phi(\mu_i), \, \text{with } \mu_i = \frac{d_+(\xV_i) - d_-(\xV_i)}{d_+(\xV_i) + d_-(\xV_i)}\, .
\end{equation}
The argument $\mu_i$ is based on the difference between the distance $d_+(\xV_i)$ from its position to the closest prototype with the same label and the distance $d_-(\xV_i)$ to the closest prototype with a different label, normalized to the range $\mu_i \in [-1,1]$.
The function $\Phi(\cdot)$ is monotonically increasing and is usually chosen to be identity $\Phi(x)=x$ or the logistic function $\Phi(x) = 1/(1+\exp(-kx))$.
The above cost function can be minimized with respect to the prototypes $\mathbf{w}$ and the relevance matrix $\mathbf{\Omega}$ by either batch- or stochastic gradient descent.
To formulate the update rule with respect to $\mathbf{w}_\pm$ and $\mathbf{\Omega}$ for the example $\xV_i$, one applies the chain rule:
\begin{equation} \label{eq:updateRule}
    \mathbf{w}_\pm = \mathbf{w}_\pm - \alpha \Phi'(\mu_i) \frac{\partial \mu_i}{\partial d_\pm} \frac{\partial d_\pm}{\partial \mathbf{w}_\pm}\, , \quad \mathbf{\Omega} = \mathbf{\Omega} - \beta \Phi'(\mu_i) \frac{\partial \mu_i}{\partial d} \frac{\partial d}{\partial \mathbf{\Omega}}
\end{equation}
In case $\Phi(\cdot)$ is a zero-centered sigmoidal function, the largest updates occur for examples close to the decision boundary with $\mu_i \approx 0$, which causes the classifier to learn faster from the most informative examples.

\subsection{Learning rules for complex-valued data}
\label{sec:learningRule}
In the complex-valued data space $\mathbb{C}^N$, the squared distance in Eq.~\eqref{eq:eucliddist} between an object $\xV_i$ and an example $\mathbf{w}_\pm$ is always real:
It is the sum of the squared magnitudes of the components of the projected difference vector $\mathbf{\Omega}(\xV_i - \mathbf{w}_\pm)$. Therefore only the innermost derivatives of the distance measure in Eq.~\eqref{eq:updateRule} are with respect to the complex-valued variables.
These can be done in an elegant way using the Wirtinger differential operators \cite{Wirtinger1927} as proposed in \cite{Gay2016}:
\begin{equation}
    \frac{\partial}{\partial z} = \frac12 \left( \frac{\partial}{\partial x} - i \frac{\partial}{\partial y} \right)\, , \quad \frac{\partial}{\partial z^*} = \frac12 \left( \frac{\partial}{\partial x} + i \frac{\partial}{\partial y} \right)\, ,
\end{equation}
in which $z = x + iy$ and $z^* = x - iy$, the complex conjugate. Using the differential operator with respect to the conjugate of the complex variable, the inner most derivatives in Eq.~\eqref{eq:updateRule} are as follows:
\begin{equation}
    \frac{\partial d*}{\partial \mathbf{w}_\pm^*} = -\mathbf{\Omega}^H \mathbf{\Omega} (\xV_i - \mathbf{w}_\pm), \quad \frac{\partial d}{\partial \mathbf{\Omega}^*} = \mathbf{\Omega} (\xV_i - \mathbf{w}_\pm) (\xV_i - \mathbf{w}_\pm)^H\, ,
\end{equation}
which are conceptually very similar to the derivatives for real-valued variables.

\section{Experiments}

In this section, we show the effectiveness of the proposed embedding approach on a set of benchmark data typically used in the area of proximity-based supervised learning.
The following section contains a brief description of the datasets with details in the references.
Subsequently, we evaluate the performance of our embedding approach on these datasets compared to some baseline classifier.

\subsection{Datasets}
\label{sec:datasets}

We use multiple standard benchmark data for similarity-based learning. All data sets used in this experimental setup are indefinite with different spectral properties.
If the data are given as dissimilarities, a corresponding similarity matrix can be obtained by double centering \cite{Pekalska2005a}:
${S} = -{J} {D} {J}/2$ with ${J} = ({I}-\mathbf{1}\mathbf{1}^\top/N)$, 
with identity matrix ${I}$  and vector of ones $\mathbf{1}$.
% TODO Double Centering + signature doppelt - auch vorher im text
The datasets used for the experiments are described in the following and summarized in Table \ref{tab:datasets}, with details given in the references.
The triplet $(p,q,z)$ is also referred to as the signature.
In this context, the signature describes the ratio of positive to negative and zero eigenvalues of the respective data set.

\begin{table}
	
	\centering
\begin{tabular*}{1\linewidth}{@{\extracolsep{\fill}} lccc}
		\toprule	
%	\begin{tabular}{l|cc}
%	\hline
	Dataset 		                &      \#samples	&   \#classes   &   signature      \\ %&   source	\\\hline
    \hline
%    Aural Sonar			            &       $100$	    &   $2$	        &   $(62,38,0)$	        \\%&       \cite{Pekalska20091017} \\
    % Bacteria			            &       $2007$	    &   $21$	    &   $(2007,0,0)$	     \\%   &       \cite{Pekalska20091017} \\
    Balls3d			                &       $200$	    &   $2$	    &   $(48,152,0)$	   \\%     &       \cite{Pekalska20091017} \\
    Balls50d			            &       $2000$	    &   $2$	    &   $(853,1147,0)$	\\%    &       \cite{Pekalska20091017} \\
    % Caltech			                &       $8677$	    &   $21$	    &   $(2258,1899,43)$	    \\    %&       \cite{Pekalska20091017}\\
%    Catcortex			            &       $65$	    &   $4$	        &   $(49,16,0)$	    \\    %&       \cite{Pekalska20091017}\\
    Chromosomes 			            &       $4,200$	    &   $21$	    &   $(2258,1899,43)$	    \\    %&       \cite{Pekalska20091017}\\
    DelftGestures  			        &       $1,500$	    &   $20$	    &   $(963,536,1)$	    \\    %&       \cite{Pekalska20091017}\\
% %    FaceRec	 		                &       $945$	    &   $139$	    &   $(794,150,1)$	    \\    %&       \cite{Pekalska20091017}\\
%     Flowcyto-1                       &       $612$       &   $3$         &   $(538,73,1)$            \\ 
%     Flowcyto-2                       &       $612$       &   $3$         &   $(26,73,582)$           \\ 
%     Flowcyto-3                       &       $612$       &   $3$         &   $(541,70,1)$            \\ 
%     Flowcyto-4                       &       $612$       &   $3$         &   $(26,73,582)$           \\ 
%     Gauss with overlap 			    &       $1,000$	    &   $2$	        &   $(469,531,0)$	    \\    %&       \cite{Pekalska20091017}\\
%     Gauss without overlap			&       $1,000$	    &   $2$	        &   $(468,532,0)$	    \\    %&       \cite{Pekalska20091017}\\
%     Patrol			                &       $241$	    &   $21$	    &   $(233,8,0)$	    \\    %&       \cite{Pekalska20091017}\\
%     Prodom                           &       $2604$      &   $53$        &   $(1502,680,422)$          \\    %&       (blinded reference) \\% \cite{Schleif2015f}\\    
    Protein                          &       $213$       &   $4$         &   $(170,40,3)$              \\    %&       (blinded reference) \\% \cite{Schleif2015f}\\    
    Sonatas			                &       $1068$	    &   $5$	        &   $(1063,4,1)$	    \\    %&       \cite{Pekalska20091017}\\
    % SwissProt  	 		            &       $10,988$    &   $10$	    &   $(8487,2500,1)$          \\       %&       (blinded reference) \\% \cite{Schleif2015f}\\
    % Tox-21 (AllBit)    		        &       $14,484$     &   $2$	        &   $(2049,0,12435)$                        \\       %&       (blinded reference) \\% \cite{Schleif2015f}\\
    % Tox-21 (Assymetric)   	        &       $14,484$     &   $2$         &	$(1888,3407,9189)$                       \\       %&       (blinded reference) \\% \cite{Schleif2015f}\\
    % Tox-21 (Kulczynski)              &       $14,484$     &   $2$         &	$(2048,2048,10388)$                       \\       %&       (blinded reference) \\% \cite{Schleif2015f}\\
    % Tox-21 (McConnaughey)            &       $14,484$     &   $2$         &	$(2048,2048,10388)$                       \\       %&       (blinded reference) \\% \cite{Schleif2015f}\\
    % Vibrio                           &       $1100$       & $49$         &   $(851,248,1)$                        \\    %&       (blinded reference) \\% \cite{Schleif2015f}\\
    
    % Voting			                &       $435$	    &   $2$	    &   $(178,163,94)$	    \\    %&       \cite{Pekalska20091017}\\
    Zongker			                &       $2000$	    &   $10$	    &   $(1039,961,0)$	    \\    %&       \cite{Pekalska20091017}\\
    
    \hline
    \end{tabular*} 
    \caption{Overview of the datasets used in our experimental setup. Details are given in the textual description.\label{tab:datasets}}
\end{table}

\begin{enumerate}

%
% AURAL SONAR DATASET
%

% \item \textbf{Aural Sonar} consists of 100 signals with two classes, representing sonar signals dissimilarity measures to investigate the human ability to distinguish different types of sonar signals by ear. Details are provided in \cite{Chen2009901}.

%
% BALLS3D BALLS50D DATASET
%

\item \textbf{Balls3d/Balls50d} consist of 200/2000 samples in two/four classes.
The dissimilarities are generated between two constructed balls using the shortest distance on the surfaces.
The original data description is provided in \cite{DBLP:conf/sspr/PekalskaHDSB06}.

% \item The \textbf{Caltech-101} data set (Fei-Fei et al., 2004) is an object recognition benchmark data set consisting of 8677 images from 101 object categories. Similarities between images were computed using the pyramid match kernel (Grauman and Darrell, 2007) on SIFT features (Lowe, 2004). Here, the similarity is PSD.

% \item The \textbf{Catcortex} data set is provided as a 65x65 dissimilarity matrix  describing the connection strengths between 65 cortical areas of a  cat from four regions (classes): auditory (A), frontolimbic (F),  somatosensory (S) and visual (V). The dissimilarity values are measured on an ordinal scale.

\item The Copenhagen \textbf{Chromosomes} data set constitutes 4,200 human chromosomes from 21 classes represented by grey-valued images.
These are transferred to strings measuring the thickness of their silhouettes.
These strings are compared using edit distance.
Details are provided in \cite{neuhaus}.

\item The \textbf{Delft gestures} (1500 points, 20 classes, balanced, signature: (963,536,1)), taken from \cite{PrTools:2012:Online}, is a set of dissimilarities generated from a sign-language interpretation problem (see Figures 8 to 10c). It consists of 1500 points with 20 classes and 75 points per class. The gestures are measured by two video cameras observing the positions of the two hands in 75 repetitions of creating 20 different signs. The dissimilarities are computed using a dynamic time-warping procedure on the sequence of positions (Lichtenauer, Hendriks, Reinders, 2008).

\item {\bf Protein}: the Protein data set has sequence-alignment similarities for 213 proteins and is used for comparing and classifying protein sequences according to its four classes of globins: heterogeneous globin (G), hemoglobin-A (HA), hemoglobin-B (HB) and myoglobin (M). The signature is (170,40,3), where class one through four contains 72, 72, 39, and 30 points, respectively \cite{DBLP:journals/pami/HofmannB97}.

\item \textbf{Sonatas} dataset consists of 1068 sonatas from five composers (classes) from two consecutive eras of western classical music.
The musical pieces were taken from the online MIDI database \emph{Kunst der Fuge} and transformed to similarities by normalized compression distance \cite{DBLP:phd/de/Mokbel16}.

\item \textbf{Zongker} dataset is a digit dissimilarity dataset.
The dissimilarity measure was computed between 2000 handwritten digits in 10 classes, with 200 entries in each class \cite{Jain19971386}.

\end{enumerate}

\subsection{Results}

In this section, we evaluate the performance of the proposed embedding on the mentioned datasets using a model that is able to handle complex-valued data.
For this purpose, we use the Generalized Learning Vector Quantization (GLVQ) from Sec. \ref{sec:cglvq} with the learning rule for complex-valued data from Sec. \ref{sec:learningRule}.
The GLVQ was parametrized once with and once without relevance learning.
%To evaluate the accuracy results of the complex-valued GLVQ, we also applied a nearest neighbor as baseline classifier
Within the GLVQ, we verified that the corrected input matrix was indeed psd by an additional test using an eigendecomposition, no fails were found.
For the initialization of prototypes, we used one prototype per class for the G(M)LVQ.
In the embedding step of Algorithm \ref{alg1} we set the meta parameter $m$ (\# of landmarks) by a rule of thumb:
\begin{itemize}
    \item If the number of data points in the data set is $< 1000$, our recommendation is $m = 40$.
    \item For data sets with a size $1000 < N < 5000$, we recommend $m = 70$.
    \item otherwise $m = 100$.
\end{itemize}

Additionally, to the GLVQ, we chose a nearest neighbor classifier (\emph{NN}) for performance comparisons to our proposed method, which remains valid also for generic psd data.
In the nearest neighbor, the kernel matrix was left in the original - that means in the uncorrected, indefinite - state.

Experiments were run in a ten-fold cross-validation. Mean prediction accuracy on the hold out test data and the respective standard deviation is reported and shown in Table \ref{tab:results}.
% TODO -- verstehe ich nicht?
%Experiments were repeated at least three times for each benchmark dataset in order to ensure the validity and the reproducibility of the results.

\begin{table}[]
% \begin{tabular}{l|c|c|c}
\begin{tabular*}{1\linewidth}{@{\extracolsep{\fill}} lccc}

\toprule
Dataset       & CGLVQ               & CGMLVQ           & Nearest Neighbor          \\
\midrule
Balls3d       & $0.61 \pm 0.07$     & $\mathbf{0.99  \pm 0.03}$     & $0.48 \pm 0.07$           \\
Balls50d      & $0.29 \pm 0.04$     & $\mathbf{0.50  \pm 0.06}$     & $0.25 \pm 0.02$           \\
Chromosomes   & $0.92 \pm 0.01$     & $0.94  \pm 0.01$              & $\mathbf{0.95 \pm 0.02}$  \\
DelftGestures & $0.94 \pm 0.02$     & $\mathbf{0.96  \pm 0.02}$     & $0.95 \pm 0.01$           \\
Protein       & $0.91 \pm 0.07$     & $\mathbf{0.98  \pm 0.05}$     & $0.22 \pm 0.04$           \\
Sonatas       & $0.82 \pm 0.03$     & $0.89  \pm 0.03$              & $\mathbf{0.90 \pm 0.01}$  \\
Zongker       & $0.87 \pm 0.03$     & $\mathbf{0.92 \pm 0.02}$      & $0.58 \pm 0.05$           \\
\bottomrule
\end{tabular*}
\caption{Prediction accuracy (mean $\pm$ standard-deviation) for the cGLVQ variants and the nearest neighbor classifier. Column \emph{cGLVQ} shows the performance of the Generalized Learning Vector Quantization without relevance learning and column \emph{cGMLVQ} provides the performance of the complex-valued Generalized Matrix Learning Vector Quantization,
employing relevance learning. Column \textit{Nearest Neighbor} shows the performance of the baseline classification.\label{tab:results}}
%\end{tabular}
\end{table}
If the data were left uncorrected we obtained often a rather poor result using the nearest neighbor classifier, sometimes even significantly worse compared to cGLVQ and cGMLVQ (see balls3d, balls50d, protein, zongker). 
In some cases, NN had equal or slightly better performance than the two cG(M)LVQ variants (Chromosomes, Sonatas).
This is due to the spectrum of eigenvalues: Chromosomes has many eigenvalues, which are almost negligible and close to zero.
Sonatas has only a few negative eigenvalues and these eigenvalues are also close to zero.
With respect to the two cGLVQ variants, one can also see a difference in performance between GLVQ and GMLVQ: the activation of relevance learning within the cGMLVQ leads to significantly better results in some cases.
However, even the mere use of the cGLVQ without relevance learning leads to a significant increase compared to the NN with uncorrected data.
Therefore, we assume that a correction step, like our embedding approach, is indeed essential since the use of uncorrected non-psd data shows a clear drop in accuracy.

In summary, the presented approach, applying an embedding of the indefinite input data into a complex-valued vector space, shows promising results on a variety of data sets.

\section{Conclusions}
In this work, we proposed a complex-valued embedding and processing pipeline to analyze non-metric or non-psd proximity data.
The approach shows very promising performance on a variety of datasets and is easy to employ.
A careful combination of approximation techniques, derived by the authors in former work, permits a valid and still effective calculation of the embedding matrix. By processing the embedding matrix, a straight forward out of sample extension is obtained, which is not easily available for many traditional indefinite learning approaches.
The low-rank embedding has the benefit that the reconstructed matrix approximates the original indefinite kernel with low error, hence all major information in the original data is preserved.
In the classifier models, a norm operator is used which eventually leads to a psd dissimilarity representation.
The effect of this operation, which is not equivalent to a classical flip correction on the original indefinite kernel, has to be evaluated in more detail in future work.
% TODO - das muesste man sich eigentlich mal genauer ansehen
% Das complex encoding (in der approximation) macht zumindest was mit der Datendarstellung auch wenn der original kernel reconstruierbar bleibt, insofern
% ist das auch mit norm nicht identisch zu einem flip auf den ursprungsdaten
% Wäre aber die frage was es dann ist ;-)
Using learning algorithms for complex-valued embeddings, predictive models can be obtained with low computational costs.
In this initial work, we focused on complex-valued G(M)LVQ and Nearest Neighbor to calculate classification model, but this will be extended to further modeling approaches in future work.
Our initial findings show that the suggested complex-valued embedding of indefinite proximity data, combined with complex-valued classifier models is a promising
very effective approach to more complicated alternatives.
%% TODO 

\bibliographystyle{splncs04}
%% Literaturverzeichnis hier erzeugen
% \bibliography{pub_fms,indefinite_kernel,non_metric,simbad,cGMLVQ,addon}
% arxiv-only with main.bib-file
\bibliography{main}

\end{document}

%% file: 01_Preamble/CommonPackages.tex
\usepackage[utf8]{inputenc}

\usepackage[T1]{fontenc}

\usepackage[english]{babel} 

\usepackage{calc}

\usepackage{graphicx}

\usepackage{float}

\usepackage[table]{xcolor} 

\usepackage{booktabs}

\usepackage{ragged2e}

\usepackage{amsmath}
\usepackage{mathtools}
\usepackage{esvect}
\usepackage{amssymb}
\usepackage{clrscode3e}

\usepackage{caption}
\usepackage{subcaption}

\usepackage{url}

\usepackage{enumitem}%

\usepackage{pgfplots}
\pgfplotsset{compat=1.11}

\usepackage{courier}
\usepackage{listings}
\usepackage{color}